# Model based neuro-fuzzy ASR on Texas processor

**Hesam Ekhtiyar[1], Mehdi Sheida[2], Somaye Sobati Moghadam[3]**

**[1]Department of Electrical and Computer Engineering,
Hakim Sabzevari University, Sabzevar, Iran.**
**[2]Department of Electrical and Computer Engineering,
Hakim Sabzevari University, Sabzevar, Iran.**
**[3] Department of Electrical and Computer Engineering,
Hakim Sabzevari University, Sabzevar, Iran.**

### Abstract
In this paper an algorithm for recognizing speech has been proposed. The recognized speech is used to execute related commands which use the MFCC and two kind of classifiers, first one uses MLP and second one uses fuzzy inference system as a classifier. The experimental results demonstrate the high gain and efficiency of the proposed algorithm. We have implemented this system based on graphical design and tested on a fix point digital signal processor (DSP) of 600 MHz, with reference DM6437-EVM of Texas instrument.

## 1.Introduction
Speech processing is divided into two sections: speaker recognition and speech recognition. Speaker recognition objective is to identify the identity of the speaker and speech recognition can convert speech to: text, to be commanded and etc. Usual speech recognition systems use separated voice commands which have a silence zone between each vocal command [1]. These vocal commands can be extracted by a simple (Voice activity detection) VAD algorithm. On the other hand there are intelligent algorithms such as SNR and Clustering, which have the ability to distinguish speech from background noise.

In order to recognize speech, specific features must be extracted from the voice. Since this signal is transient at the time domain, it is usually transferred to the frequency domain for further calculations. This can be done using the Discrete Fourier Transform (DFT) or the wavelet transform. Then filters that amplify the main signal and weaken the signals noise are used. The most common is using the MFCCs [2] which act as humans ears. At the end an algorithm for matching speech with its relevant class is used such as DTW or neural networks. Fig. 1 illustrates the proposed methods process as a fellow chart.

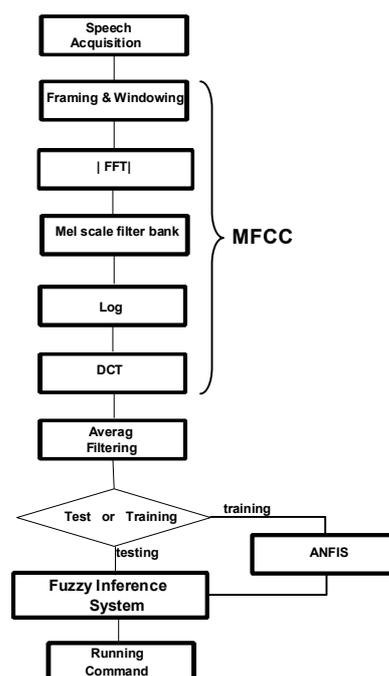

Fig. 1 The flow chart of the proposed algorithm's process.

Researchers always want to design easily and faster and also do not face details of implementation such as working with pointers in C/C++ and etc [3,4]. in the [5] a real time ASR system is implemented on a float point DSP by C programming in CCS environment and in [6] an FPGA based speech recognition has been implemented in automotive environment. One of the best approaches to design systems fast and easily is model base design such as simulink/matlab [3], labview etc. in this way the rules of software engineering are usually observe automatically and also it has a lot of advantages [7] in other hand, simulink has a code generation especially for Texas instrument processors [7]. DM6437 is a fix point DSP and compatible with simulink and also is faster than floating point DSPs [8,9,10] which is shown in fig. 2.

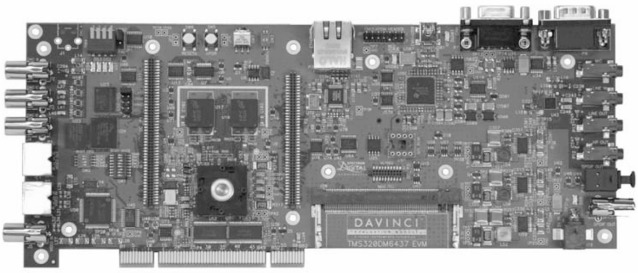

Fig. 2 DSP board (DM6437).

## 2. Our Approach

### 2.1 Database

A database has been created from 4 commands: left, right, up and down commands which are recorded from two people in different conditions both having a 30cm distance from the microphone. The recorded speech had a .wav format extension with specifications as table 1:

Table 1: Recorded wav files specifications

| Specifications | Value |
|---|---|
| Sampling Freq. | 48kHz |
| Sampling size | 16bit |
| No. of channels | mono channel |
| Bit rate | 96kbps |

The total number available speech commands are 96. A subsample of 48 speech commands were separated randomly for training process and the other 48 were used for the testing process.

### 2.2 DFT

DFT is used to transfer a signal to frequency domain discretely. After the signal is split into frames and windowing, the FFT of each frame is calculated and placed in continuous columns. The horizontal axis of the output matrix indicates time and the vertical axis shows frequency. The value of each place shows the power or amplitude of the signal [11].

### 2.3 MFCC

The main idea of this part is to extract a unique feature vector for each vocal command. This command can be recognized using the LPC analyzer [12], the DFT analyzer and the most common analyzer: the MFCC [2,13].Fig.1 show this algorithm which containing the framing block to DCT block.

The DCT [14] was calculated as:

$$C_K = \sum_{i=1}^{N} \log(E_i) \times \cos\left[\frac{\pi k}{N}\left(i - \frac{1}{2}\right)\right] \quad (1)$$

### 2.4 ANFIS

Previous speech recognition methods are based on the HMM, DTW and the ANN which have their own advantages and disadvantages. Neural networks are one of the strongest speech recognition methods which have a magnificent growth in recent years [15]. Fuzzy logic which proposed by Professor Zadeh in 1960 assisted neural networks and these two fields, together have joined to create a powerful tool named ANFIS [16].Adaptive Neuro-Fuzzy Inference Systems (ANFIS) is one of the most powerful speech recognition tools. ANFIS is narrowed to Sugeno system and is shown in Fig. 3. The relations between the layers are discussed extensively in [17].

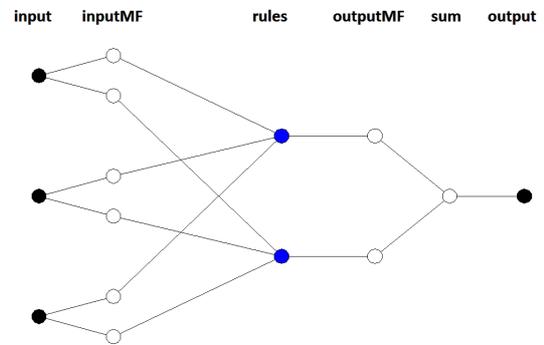

Fig. 3 The structure of ANFIS.

Neuro-Fuzzy networks use two learning methods: the back propagation method and the hybrid method. Which the hybrid method has a higher convergence speed [17].

### 2.5 Feature Compression

A new simple method is used to compress MFCC features from voice commands which have two main characteristics, in comparisons with other methods it has better output and speed up training phase because of very small size of feature vector.

Thirteen Cepstral coefficients are extracted from each vocal command [14]. The 13xN matrix is transformed to the final feature vector as:

$$feature\ vector(i) = \frac{1}{N}\sum_{i=1}^{N} cepstral(i, j) \quad (2)$$

For example Fig. 4. for two commands "right" and "left" shows the features that were extracted.

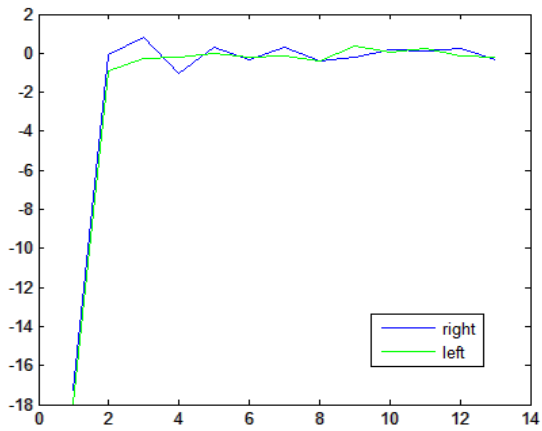

Fig. 4 The feature vectors for the 'right' and 'left' commands.

In our proposed algorithm a fuzzy neural network has been used which has the specific characteristic. its specification is shown in table 2 and is illustrated in Fig. 5.

Table 2: Neuro-fuzzy networks specifications

| Feature | Value |
|---------|-------|
| No. of Inputs by Outputs | 13x1 |
| Networks structure | subtractive clustering |
| Cluster center's range of influence | 0.2 |
| No. of rules | 48 |
| Learning regulation | Hybrid |
| Input membership functions | Gaussian MF |

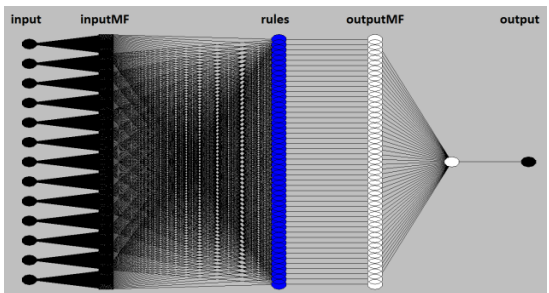

Fig. 5 The used fuzzy neural network.

As mentioned before, we randomly selected half of the data base utterances for training and others for testing in step 1 and in step 2, utterances is exchanged and recognition rate has shown in table 3.

Table 3: Recognition rate of 48 test utterance

| | Step 1 | | Step 2 | |
|---|---|---|---|---|
| | Speaker 1 | Speaker 2 | Speaker 1 | Speaker 2 |
| MLP | % 100 | % 98 | % 100 | % 95 |
| ANFIS | % 100 | % 100 | % 100 | % 100 |

## 3. Evaluate on the Hardware

As mentioned earlier, Simulink uses a block based approach to algorithm design and implementation [7,18]. Fig. 6 has shown our algorithm which has been designed in simulink for evaluation in PC.

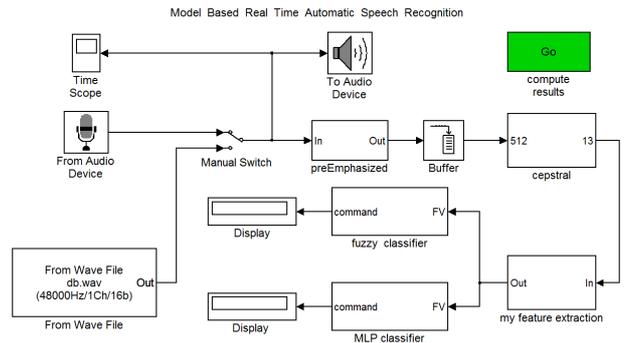

Fig. 6 The model of implemented algorithm in PC.

As you can see in fig. 6, it can acquire signal from input devices such as microphone, mp3 player etc and after cepstrals computation, our feature extraction algorithm extract unique feature vector for one utterance and show related command after classifying by MLP and neuro-fuzzy network simultaneously in displayer's.

A neuro-fuzzy network has been evaluated on DM6437 DSP board which is shown in fig8. In our database four directions ("left", "right", " top" and "down") was considered and the results is shown in table3. In [12] an LPC and MLP based speech recognition has been proposed on 1GHz DSP. Main programming environment for Texas processors is code composer studio (CCS) for writing C, C++ and Asembly code [9,19] and there is a program in it for connects to matlab which called MATLAB Link for Code Composer Studio [20]. It helps to use matlab functions to communicate with Code Composer Studio and with information stored in memory and registers on a DSP. With the links, information could be transferred to and from Code Composer Studio with a communication between CCS and DSP through the RTDX. Real-Time Data Exchange provides real-time, continuous visibility into the way target board applications operate in the real world. RTDX allows transferring data between a host computer and target devices without interfering with the target application.

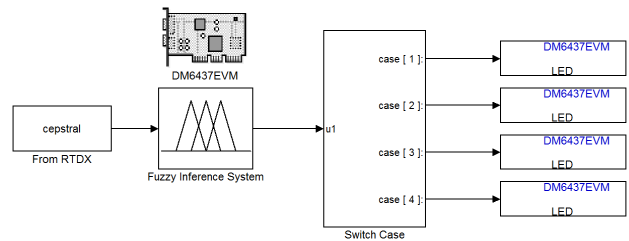

Fig. 7 Tested model on DSP.

## 4. Experimental result

Neural networks have a good performance, if input data is normalized and is in a limited range, but in the neuro-fuzzy networks there is not any restriction in input feature vector.For this reason, the first channel of cepstral coefficient which contain of dc frequency has been eliminated in MLP training phase for improving performance, but still our algorithm has better result.

# 5. Conclusion

In this paper an automatic speech recognition system was presented which introduce a faster and easier method . In our approach MFCC is used as a feature extractor and neuro-fuzzy network as a classifier based on a DSP board. In comparison with MLP our algorithm improved performance .

## Acknowledgment

We would like to thank Hakim Sabzevari University for them invaluable support through this work. The research presented in this paper is based upon work supported by, or in part by, a research grant.

**Hesam Ekhtiyar** received the B.S. degree in computer engineering from Hakim Sabzevari University, Sabzevar, Iran, in 2011. his research interests include computer vision, speech recognition, robotics, soft computing. hekhtiyar@gmail.com

**Mahdi Sheida** received the B.S. degree in computer engineering from Hakim Sabzevari University, Sabzevar, Iran, in 2011. his research interests include computer vision, speech recognition, network programming. m.sheida87@gmail.com

**Somayeh Sobati Moghadam** is lecturer in Hakim Sabzevari University. She received her B.Sc. degree in Applied Mathematics in computer in 2001 from Amir Kabir University of technology and M.Sc. degree in IT in 2007 from INSA-Lyon- France Univercity. From 2010 she is lecturer in Hakim Sabzevari University.Her research interests include Computer Vision, GIS and Security. s.sobati@sttu.ac.ir